# GPU Accelerated Color Correction and Frame Warping for Real-time Video Stitching

**Lu Yang** · **Zhenglun Kong** · **Ting Li** · **Xinyi Bai** · **Zhiye Lin** · **Hong Cheng**

Center for Robotics, University of Electronic Science and Technology of China, Chengdu, China
kong.zhe@northeastern.edu, wwwbxy123@163.com, 940872387@qq.com, {yanglu, hcheng}@uestc.edu.cn

**Abstract** Traditional image stitching focuses on a single panorama frame without considering the spatial-temporal consistency in videos. The straightforward image stitching approach will cause temporal flicking and color inconstancy when it is applied to the video stitching task. Besides, inaccurate camera parameters will cause artifacts in the image warping. In this paper, we propose a real-time system to stitch multiple video sequences into a panoramic video, which is based on GPU accelerated color correction and frame warping without accurate camera parameters. We extend the traditional 2D-Matrix (2D-M) color correction approach and a present spatio-temporal 3D-Matrix (3D-M) color correction method for the overlap local regions with online color balancing using a piecewise function on global frames. Furthermore, we use pairwise homography matrices given by coarse camera calibration for global warping followed by accurate local warping based on the optical flow. Experimental results show that our system can generate high-quality panorama videos in real time.

**Keyword**   Video Stitching · Frame Warping · Spatio-temporal Color Correction · Multi-camera · GPU acceleration

## 1 Introduction

While image stitching has been well studied [33] as the process to seamlessly merge the input images which have limited overlap regions into a panoramic image with wide view field [48], video stitching is much more challenging due to the requirements of temporal smoothness and online computation. Real-time video stitching systems are important for many emerging applications such as teleconference, human-machine interaction, virtual reality and large-scale video surveillance [45, 46].

There are three main steps in the process of images or video stitching [16, 34] : Geometric warping, color correction and finally panorama blending. Geometric warping projects the source images into a specified plane using the homographic matrix which is called the pre-warp. Then, the optical flow estimation is used to warp overlap regions accurately which is called local warping [22, 41, 2, 51]. Color correction removes all the color differences caused by different views and different exposure levels under different situations. Finally, image blending stitches images with limited overlapped regions into a panoramic image. A lot of work has been done in the domains of image matching and blending, while geometric warping and color correction have received less attention from relatively simpler treatment [43]. Recently, with the growing demand and popularity of high definition images and videos, people have begun to recognize the importance of geometric warping and color correction [31, 17, 18].

Generally, color correction schemes try to match the color of each image in the panorama to an arbitrarily selected reference image in order to solve the unexpected color differences between neighboring views which arise due to different exposure levels and view angles [15, 27]. Color correction works [43] can be divided into parametric and non-parametric approaches. Parametric approaches assume a relation between the colors of the target image and those of the source image, whereas non-parametric approaches do not follow any particular model and most of them use some form of a look-up table to record the mapping for the full range of color intensity levels. Although parametric approaches are generally better than non-parametric approaches in extend-

Lu Yang
Center for Robotics, School of Automation Engineering, University of Electronic Science and Technology of China, Chengdu 611731, Sichuan, China
Tel.: +86 028-61830797
Fax: +86 028-61830797
E-mail: yanglu@uestc.edu.cn

ability, they do not use spatial and temporal information. Texture is also considered to be a way to improve image and video stitching, Yongsheng et al. proposed a multi-scale rotation-invariant representation (MRIR) of textures by using multiscale sampling [44]. In this paper, we aim to provide an spatio-temporal correction algorithm which can handle videos or image sequences in real time system.

Our spatio-temporal color correction approach generally includes 3D-M color transfer and online color balancing based on the piecewise function. In the computer vision and multi-view video processing communities, the efforts on solving the color balancing problem for multi-view stitching used exposure compensation (or gain compensation). Uyttendaele [36] smoothly interpolated parameters which computed exposure corrections on a block-by-block basis using a spline to get spatially continuous exposure adjustment. Nanda and Cutler [28] first incorporated gain adjustment as part of the AutoBrightness function for their multi-head panoramic camera called RingCam. Later, Brown and Lowe [8] employed it in their well-known automatic panorama stitching software "Autostitch". Tian [35] used histogram mapping over the overlap areas to estimate the transformation matrix $M$. Zhang's work [49] was based on the principal regions mapping to estimate $M$ where the highest peaks in the hue histogram were designated as principal regions. Reinhard [30] proposed a linear transformation based on the simplest statistics of global color distributions of two images. When the video frames contain many pedestrians, hierarchical feature selection[39, 40] or spectral embedded clustering can be used[37, 38]. In this paper, we propose a novel algorithm of color balancing for real-time video stitching system along with the color transfer based on Tian's work to remove color differences between images.

On the other hand, geometric warping includes two procedures. The first procedure named pre-warp obtains the pair-wise homographic matrices between images through internal and external camera parameters. Then, the homographic matrices are used to warp images into the same coordinate. Accurate camera parameters are important in capturing image and videos. Camera calibration is the process of estimating the parameters of the camera using images of a special calibration pattern. The parameters include camera intrinsic, distortion coefficients, and camera extrinsic. No high-accuracy calibration means no high-accuracy measurement. Images and videos taken by cameras with inaccurate camera parameters will cause distortion in the edge of the image. So, it will be hard to stitch two or more images/videos together. The second procedure named local warping is a fusion method to guarantee the optimal match point of overlap regions through optical flow estimation [9]. Common methods of warping produce ghost in the overlap regions caused by large displacements and parallaxes as shown in figure 1. Pre-warp generally uses calibration to calculate homographic matrices [19]. Classical calibration methods are based on specific experimental conditions such as the Black and white checkerboard to obtain accurate and robust calibrating results [23, 32, 50]. Therefore, they are still the dominant calibration approaches. However, it would be no longer valid in many on-line computer vision or photogrammetry tasks. This is because the camera intrinsic and extrinsic parameters such as focal length and camera pose can be changed intentionally in order to perform some specific operations. Besides, the calibration system is subject to accidental changes of the camera parameters due to thermal and mechanical influences. Hence, classical calibration methods are not appropriate to online recalibration tasks. By contrast, auto-calibration methods use projective constraints between images to establish the coordinate correspondence without using any calibration pattern [1, 26, 21]. Although auto-calibration methods are very convenient for calibration tasks, the state-of-the-art auto-calibration approaches are not robust enough to estimate accurate parameters and recalibrating may be required [33]. In this paper, a novel method is proposed to calculate new homographic matrices accurately based on SURF algorithm on blending regions which apply to the erroneous parameters and can robustly stitch images without the checkerboard.

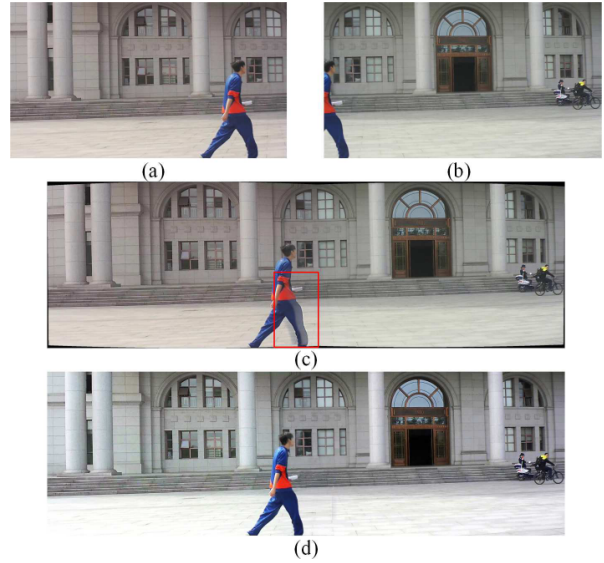

**Fig. 1** The comparison of stitched images. Image (a) is the left image and (b) is the right image. Image (c) is the panorama image using SIFT feature matching [8] which contains the ghost highlighted in the bounding box. Image (d) is our method which stitches a panorama image without the ghost.

Our video stitching system mainly involves two modules to enhance the quality of stitching. One is the spatio-temporal color correction, the other one is frame warping. The whole system uses GPU acceleration for the speed-up. Similar approaches have been done by Chengyao et al. [12]

and M. U. Aziz et al. [3]. The rest of the paper is organized as follows. Section 2 describes the implementation of the online color correction algorithm both for the overlap regions and the whole images. Section 3 presents the procedures of the auto-calibration algorithm using SURF to construct the refined homographic matrixes. Section 4 evaluates our GPU acceleration procedures. Section 5 shows the experimental comparisons between our results and the benchmark results. Section 6 concludes the paper.

## 2 Spatio-temporal color correction

In automatic multi-view video stitching, color correction is the process of correcting the color differences between neighboring views which come from different exposure and view angles [43]. Color balancing is the global correction of the intensities of stitched frames. Color transfer aims to build the relation that adjusts the color of the source image to the color of the target image in overlaps. In our system, video cameras are fixed thus color changes according to the object motion or the occlusion. Therefore, the online color correction algorithm is required to deal with sudden changes of color in the temporal domain.

### 2.1 3D-M color transfer

Our 3D-M color transfer approach uses time information of histogram to build a relation between the overlaps of images. The pipeline is as follows:

Step 1: Calculate and obtain the overlap region in the two frames using the homographic matrix $H'_{ij}$ given by geometric calibration.

Step 2: Calculate the histogram $G_i$, $G_j$ of the overlap region in frame $i$ and $j$.

Step 3: Because of the occlusion and the change of illumination, the overlap region of the two images has chromatic aberration. To deal with it, we use histogram specification to effectively control the distribution of the color histogram, and equalize the color. Most of the noisy points can be excluded and more details of the target can be preserved. First, we do histogram specification between histogram $G_i$, $G_j$ to construct a canonical histogram $G'_i$ (image $j$ is the reference map). Then, we use $G'_i$ to revise the overlap region in image $i$. The process is shown in figure 2.

Step 4: Calculate $M$ matrix.
The transform matrix $M$ from image $i$ to image $j$ can be presented as[35]

$$M = \left[ I_i^T I_i \right]^{-1} I_i^T I_j \qquad (1)$$

where $I_i$ is a $\begin{bmatrix} n & 3 \end{bmatrix}$ matrix which means the pixel value of image $i$, $I_j$ is a $\begin{bmatrix} n & 3 \end{bmatrix}$ matrix of the revised image $i$ through histogram specification in step 3, and $n$ is the number of pixels in the images.

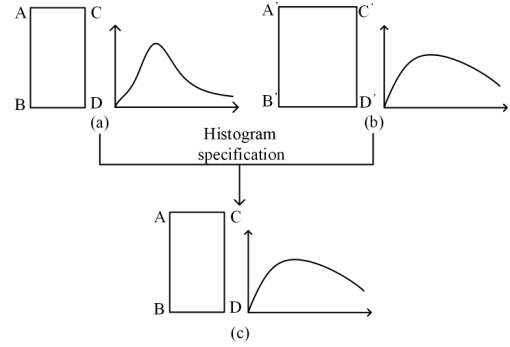

**Fig. 2** The process of histogram specification. This process inputs the overlap region of source image $i$ and target image $j$, and outputs a transfer image which has the canonical histogram of target image. Figure (a) represents the overlap region of source image $i$ and its histogram, and figure (b) represents the overlap region of source image $j$ and its histogram. Figure (c) shows the transfer image.

According to the temporal smoothness, we can assume that the information in the present frame is largely related to the information in its previous 2 frames. Hence, the time sequences should be considered in the 3D $M$ matrix. In this paper, we denote $T$ as the present frame, and $T-1$, $T-2$ as the previous 2 frames.

Next, we reconstruct the $I_i$ and $I_j$ as:

$$\begin{aligned} I_i &= \begin{bmatrix} I_{Ti} & I_{(T-1)i} & I_{(T-2)i} \end{bmatrix}^T, \\ I_j &= \begin{bmatrix} I_{Tj} & I_{(T-1)j} & I_{(T-2)j} \end{bmatrix}^T \end{aligned} \qquad (2)$$

where $I_i, I_j$ is a $\begin{bmatrix} 3n & 3 \end{bmatrix}$ matrix, $I_{Ti}, I_{Tj}, I_{(T-1)i}, I_{(T-1)j}, I_{(T-2)i}, I_{(T-2)j}$ is a $\begin{bmatrix} n & 3 \end{bmatrix}$ matrix and $n$ is the number of pixels in the images.

Finally, we reconstruct the $M$ matrix with time sequence as

$$\begin{aligned} M = &\left[ \begin{bmatrix} I_{Ti} & I_{(T-1)i} & I_{(T-2)i} \end{bmatrix} \begin{bmatrix} I_{Ti} & I_{(T-1)i} & I_{(T-2)i} \end{bmatrix}^T \right]^{-1} \\ &\cdot \begin{bmatrix} I_{Ti} & I_{(T-1)i} & I_{(T-2)i} \end{bmatrix} \begin{bmatrix} I_{Tj} & I_{(T-1)j} & I_{(T-2)j} \end{bmatrix}^T \end{aligned} \qquad (3)$$

Step 5: Use the transformed matrix $M$ to correct the overlap region of image $i$.

$$I'_i = I_i \cdot M \qquad (4)$$

where $I'_i$ is the corrected $\begin{bmatrix} n & 3 \end{bmatrix}$ matrix of the image $i$, $n$ is the number of pixels in the images.

The $M$ matrix calculated by our method is robust for color change caused by scene mutation, illumination mutation and occlusion.

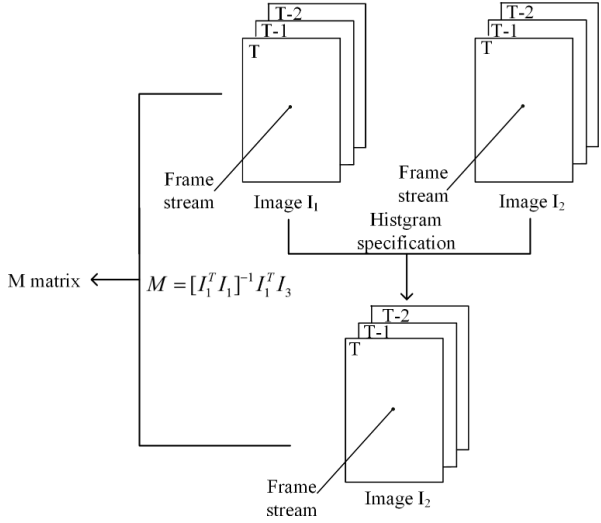

**Fig. 3** The process of histogram specification. This process inputs the overlap region of source image $i$ and target image $i$, and outputs a transfer image which has the canonical histogram of target image.

2.2 Online color balancing based piecewise function

We have implemented color transfer in the overlap region of images, we further balance the color of whole stitched image. The online color balancing instructs as follows.

The relation between the pixel value of each channel $R, G, B$ on the source image $i(x,y)$ at any given point and the pixel value in the balanced image $I(x,y)$ can be expressed as

$$I(x,y) = F(i(x,y)) \tag{5}$$

where $F(x) = a + bx$ means linear color balancing, and $F(x) = a + T(x/T)^b$ means color balancing with $\gamma$ function.

Generally, single function $F(x)$ is used for color balancing. Although it is useful in some cases, it makes corrected images more bright or dismal than the standard level. A way to solve this problem is to divide color pixels into three parts: dark, normal, and bright. The forms of function $F$ are respectively $\gamma$ color balancing, liner color correction and $\gamma$ color balancing.

The most important step is how to divide color pixels into parts. A valid scheme is using image histogram to determine the three parts of images. We assume the histogram of a source image $I_1$ is computed and equalized in each channel and choose the pixel value $m_1$ which has $\lambda$ percent proportion of the total histogram, and the pixel value $m_2$ which has $(1-\lambda)$ percent proportion. Where parameter $\lambda$ stands for critical point to black region of the source image while parameter $(1-\lambda)$ stands for the white region. In our system, we assign $\lambda$ as a value of 0.05. Then, with the certain $m_1$ and $m_2$, a relationship between the pixels of the source image and the corrected image is determined. We use liner color balancing in range $[m_1, m_2]$, and $\gamma$ color balancing in range $[0, m_1]$ and $[m_2, 255]$ (as shown in figure 4).

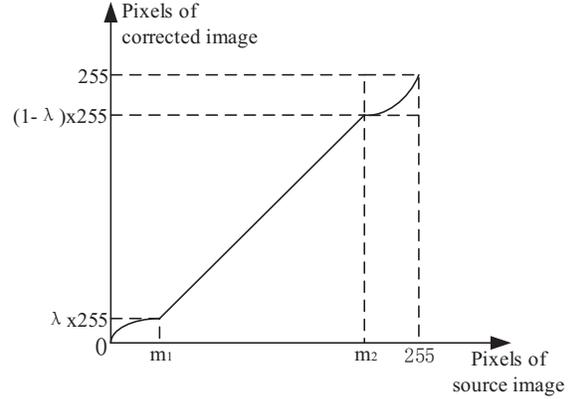

**Fig. 4** Pixel values transformation function. We use RGB image histograms to construct the distribution function which can correct the stitched images online.

To realize online color balancing, we define the threshold pixel values $m_1$ and $m_2$ in the current frame as the average value among 3 frames which contain the current frame and its last 2 frames. Thus, the threshold values cover information between 3 frames to ensure the color changes smoothly and fluently from frames.

Supposed the current frame is $T(T \geq 2)$, $T-1$ and $T-2$ are the last 2 frames. The threshold pixel values are $m_1(x)$ and $m_2(x)$, where $x$ represents frames. The $m_1(x)$ and $m_2(x)$ of frame $T(T \geq 2)$ can be expressed as

$$m_1(T) = \frac{\sum_{T-2}^{T} m_1(x)}{3} \tag{6}$$

and

$$m_2(T) = \frac{\sum_{T-2}^{T} m_2(x)}{3} \tag{7}$$

In exception, for the first frame, $m_1$ is only determined by the current frame, while for the second frame, $m_1$ is the average value of the first frame's and the current frame's. Then using $m_1$ and $m_2$ to realize online color balancing.

## 3 Coarse and fine frame warping

To find the corresponding points between two images in their overlap region, optical flow algorithm is required. The accuracy of feature point matching is directly determined by the selection of overlap region. Here the overlap region is computed by the homographic matrix which is constructed through the camera calibration. Thus, the accuracy of the homographic matrix, which acted as the chief factor in our stitching system can be guaranteed.

## 3.1 Pre-warping using coarse camera calibration

In projective geometry, a homography is an isomorphism of projective spaces, induced by an isomorphism of the vector spaces from which the projective spaces derive from. It is a bijection that maps line to line, and it is called a collineation. In general, some collineations are not homographies, but the fundamental theorem of projective geometry asserts that it is not true in the case of real projective spaces in at least two dimensions.

With the estimated parameters length, angle, position, and distance, we construct a spatial global homographic matrix $H_{ij}$ towards the plane of image $i$ to image $j$. We first project the image coordinates of the image $i$ and image $j$ to the corresponding world coordinates, then we compute the homographic matrix $H_i$ and $H_j$.

The relationship between the world coordinate system and the image coordinate system is

$$\begin{bmatrix} u \\ v \\ 1 \end{bmatrix} = \frac{1}{Z_c} K \begin{bmatrix} R_{3\times 3} & t_{3\times 3} \\ 0^T & 1 \end{bmatrix} \begin{bmatrix} X_W \\ Y_W \\ Z_W \\ 1 \end{bmatrix} \quad (8)$$

where $\begin{bmatrix} u & v & 1 \end{bmatrix}^T$ is the point in the image coordinate, $\begin{bmatrix} X_W & Y_W & Z_W & 1 \end{bmatrix}^T$ is the corresponding point in the world coordinate. $K$ is the camera intrinsic parameters matrix, which can be written as $K = \begin{bmatrix} \frac{f}{dx} & 0 & \frac{c_x}{dx}+u_0 & 0 \\ 0 & \frac{f}{dy} & \frac{c_y}{dy}+u_0 & 0 \\ 0 & 0 & 1 & 0 \end{bmatrix}$.

Let

$$U = \begin{bmatrix} u & v & 1 \end{bmatrix}^T, x = \begin{bmatrix} X_W & Y_W & Z_W \end{bmatrix}^T \quad (9)$$

then

$$U = \frac{1}{Z_c} K \begin{bmatrix} R & t \end{bmatrix} \begin{bmatrix} X_W & Y_W & Z_W & 1 \end{bmatrix}^T = \frac{1}{Z_c} K(Rx+t) \quad (10)$$

so the homography matrix towards image coordinate of $i$ to the world coordinate is

$$H_i = K_i(R_i x + T_i) \quad (11)$$

Similarly, we compute the homographic matrix $H_j$ to project the coordinate of image $j$ to the world coordinate.

Finally, we construct a pairwise homographic matrix $H_{ij} = H_i \cdot H_j^{-1}$, which describes the coordinate transformation from the image $i$ to $j$.

$$H_{ij} = K_i \begin{bmatrix} R_{i(3\times 3)} & T_{i(3\times 3)} \\ 0^T & 1 \end{bmatrix} \begin{bmatrix} R_{j(3\times 3)} & T_{j(3\times 3)} \\ 0^T & 1 \end{bmatrix}^{-1} K_j^{-1} \quad (12)$$

where $H_j \cdot H_j^{-1} = I$ in equation 12.

We conduct projective transformation to pre-warp all frames of the left image into the plane of the reference map to coarse geometric alignment. Then, we do spatial local warping to make the left image and the reference map at the same height. Considering our 3 under-stitched images are taken by the camera with fixed intrinsic parameters, the length, angle, position and distance from extrinsic parameters are need to be estimated. Only depending on these parameters are not sufficient enough to construct a precise homographic matrix because the extrinsic parameters have accidental errors. In order to alleviate the instability of the computed homographic matrix caused by both camera movements and the error of parameters, our homographic matrix should be revised to enhance the accuracy of feature matching.

Here we define a scale matrix $N$ to describe the ratio of the area from the overlapping region in left image and the reference map. Then, we define a skew matrix $M$, which is set to describe the placement of overlap region in image $i$ and the corresponding region in image $j$. Here we set image $j$ as the reference map, and image $i$ as the under-corrected image. $M$, $N$ can be calculated as

$$M = \begin{bmatrix} 1 & 0 & t_x \\ 0 & 1 & t_y \\ 0 & 0 & 1 \end{bmatrix} \qquad N = \begin{bmatrix} s_x & 0 & 0 \\ 0 & s_y & 0 \\ 0 & 0 & 1 \end{bmatrix} \quad (13)$$

where $t_x$, $t_y$ are skew parameters, and $s_x$, $s_y$ are scale parameters. They are used to calculate the value of these unknown parameters with SURF feature matching. SIFT[25,24] and SURF[5,4] are the detector and descriptor for points in images where the image is transformed into coordinates, especially for deducing homography matrixes. SIFT feature descriptor is invariant to uniform scaling, orientation, and partially invariant to affine distortion and illumination changes [47]. SURF makes a copy of the original image with pyramidal gaussian or laplacian pyramid shape to obtain an image with the same size but the bandwidth of it is reduced[29]. The SURF algorithm is based on the same principle and steps as SIFT but it is more stable and fast[20]. Hence, we choose SURF for feature matching in our experiments.

Based on our new under-corrected image which has been warped through pairwise homography mentioned above, we first compute the area of the overlap region. But the overlap region is usually unreliable since it cannot include the full feature points. So we extend the overlap region as a broader overlap region to ensure every feature point is in the overlap region. Then, we get feature descriptor in the overlap region with SURF algorithm and remove the mismatched points by RANSAC[13]. Next, we match the feature points, compute appropriate skew parameters $t_x$, $t_y$ and scale parameters $s_x$, $s_y$. Finally, we build a skew matrix $M$ and a scale matrix $N$.

The revised homographic matrix $H'_{ij}$ is the product of the initial homographic matrix $H_{ij}$, scale matrix and skew matrix can be computed as

$$\begin{aligned}H'_{ij} &= M \cdot H_{ij} \cdot N \\ &= \begin{bmatrix} 1 & 0 & t_x \\ 0 & 1 & t_y \\ 0 & 0 & 1 \end{bmatrix} \cdot K_i \cdot \begin{bmatrix} R_{i(3\times 3)} & T_{i(3\times 3)} \\ 0^T & 1 \end{bmatrix} \\ &\quad \cdot \begin{bmatrix} R_{j(3\times 3)} & T_{j(3\times 3)} \\ 0^T & 1 \end{bmatrix}^{-1} \cdot K_j^{-1} \cdot \begin{bmatrix} s_x & 0 & 0 \\ 0 & s_y & 0 \\ 0 & 0 & 1 \end{bmatrix}\end{aligned} \quad (14)$$

The process to calibrate the right image is similar to the left one which is mentioned above.

Our approach aims to construct an initial homographic matrix $H'$ through some rough camera intrinsic and extrinsic parameters with errors. By implementing on SURF feature matching, an overlap region can be computed which is mainly determined by scale matrix and skew matrix. We set the product of the initial homographic matrix, scale matrix and skew matrix as the revised homographic matrix, which will accurately calibrate our images and get rid of the traditional checkerboard.

### 3.2 Fine local warping for blending

Pre-warping aims to use camera calibration to project images onto the same plane, while local warping can match images more accurately through the optical flow method on the overlap regions. Moreover, optical flow describes the pixel motion of the space moving object thus can be widely used in object detection and tracking [10]. In our video stitching system, we use optical flow between the overlap regions which can be calculated by pre-warpping.

The optical flow stores the points of their corresponding optimal matching position. Supposing that we have image $i$ and image $j$ and their overlap regions, we calculate the optical flow from image $i$ to image $j$ named $flow_{ij}$ and its reverse optical flow named $flow_{ji}$. Then, we fuse image $i$ and image $j$ into one stitched image using weighted mean fusion function $P(x,y)$ as shown in equation 15.

$$P(x,y) = \begin{cases} P_i(x,y) & (x,y) \in I_i \cap \notin I_j \\ \theta_i P_i(x,y) + \theta_j P_j(x,y) & (x,y) \in I_i \cap \in I_j \\ P_j(x,y) & (x,y) \notin I_i \cap \in I_j \end{cases} \quad (15)$$

where $(x,y)$ is the point of the stitched image, $\theta_i$ and $\theta_j$ are the weights of each image, which are related to the position of the images.

$$\begin{cases} \theta_i = \frac{i(x,y).x}{i(x,y).x + j(x,y).x} \\ \theta_j = \frac{j(x,y).x}{i(x,y).x + j(x,y).x} \end{cases} \quad (16)$$

where $i(x,y).x$ is the x coordinate value of the image $i$, and $j(x,y).x$ is the x coordinate value of the image $j$.

In the overlap region, we fuse images using function $P(x,y) = \theta_i P_i(x,y) + \theta_j P_j(x,y)$. Local warping can be conducted as

$$\begin{cases} X_i = x + \theta_i * flow_{ij}.x \\ X_j = x + \theta_j * flow_{ji}.x \\ Y_i = y + \theta_i * flow_{ij}.y \\ Y_j = y + \theta_j * flow_{ji}.y \end{cases} \quad (17)$$

where $(X_i, Y_i)$ is the new point of image $i$, and $(X_j, Y_j)$ is the new point of image $j$.

The final fusion function $P(x,y)$ is given by

$$P(x,y) = \theta_i P_i(X_i, Y_i) + \theta_j P_j(X_j, Y_j) \quad (18)$$

## 4 GPU acceleration for real-time video stitching

We ran our experiments with GPU. The platform is CUDA. Several related works have been done on GPU acceleration. For instance, Bayazit [6] used GPU implementation for real-time gesture recognition. Gobron [14] used GPU implementation for retina simulation. Cabido [11] realized the real-time region tracking using GPU. Here, we use GPU implementation for SURF feature matching.

The computational time of our method is not only proportional to the size of input images and the number of control points, but also affected by the hardware performance. Our main computing hardware devices are Intel Xeon CPU with 3.3GHz rate and NVIDIA Quadro4000 with 3GB memory. We use the OpenCV library [7] because it provides GPU models with optical warping [9] and other functions with CUDA GPU programming. The SURF feature results are shown in figure 5. The running time of CUDA-based implementation is compared to its CPU counterpart shown in table 1.

Color correction uses matrix $C$ between the source image and the target image to correct the source image. The matrix $C$ is a 256x3 matrix which stands for three channels of image where the value of pixels has 256 levels. In the GPU implementation for color correction, we first convert image types from *Mat* to *GpuMat*, then send the data to the GPU memory. In the GPU, every pixel corresponds to a kernel. We use kernel function to control every instruction operation code. The process is shown in figure 6.

Our video stitching system has four steps for panoramic image fusion: geometric warping, color correction, local warping and image fusion. Geometric warping projects images from different perspectives into a same plane. Then, color correction removes color difference and repair distortion. Local warping uses optical flow to make it more precise. Image fusion mosaics images into a panorama image. The process of the GPU acceleration in our system is shown in figure 7.

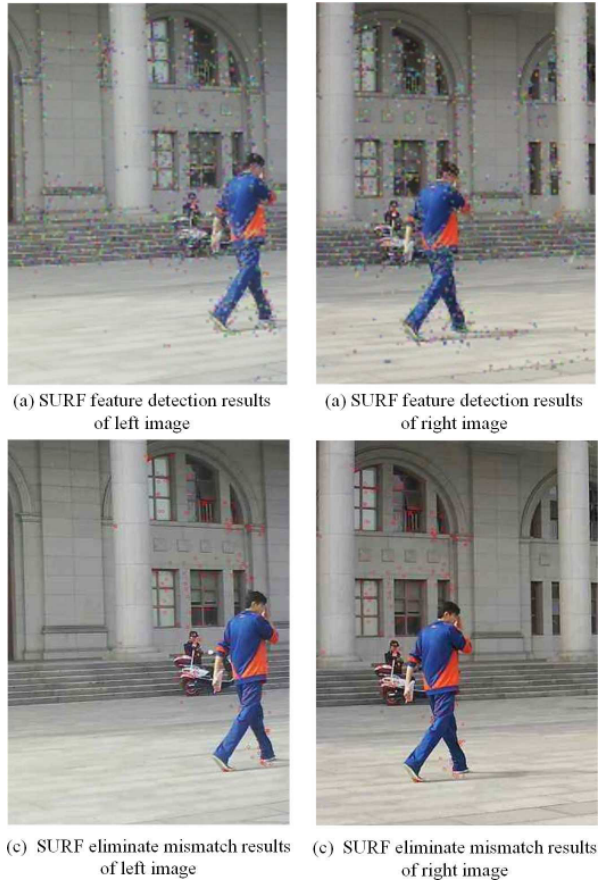

(a) SURF feature detection results of left image

(a) SURF feature detection results of right image

(c) SURF eliminate mismatch results of left image

(c) SURF eliminate mismatch results of right image

**Fig. 5** Results of SURF feature detection points. We set the overlap region of the images as input to detect SURF features and eliminate mismatch points.

**Table 1** The running times for SURF feature detection in single-thread CPU and GPU implementations. GPU can make 12 times speed-up compared to the CPU version.

| Test images | Time for SURF detection in single CPU(s) | Time for SURF detection in GPU(s) | Sum of SURF features points | Speed-Up ratio |
|---|---|---|---|---|
| Left image (789 × 1080) | 1175 | 94 | 3300 | 12.5 |
| Right image (789 × 1080) | 1199 | 94 | 3132 | 13.2 |

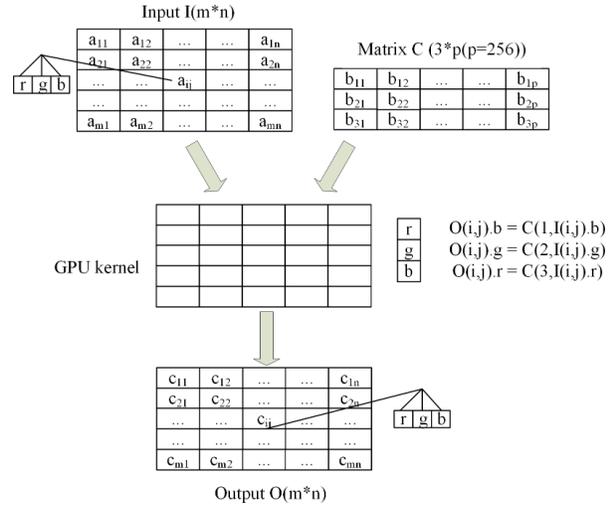

**Fig. 6** The GPU implementation of color correction.

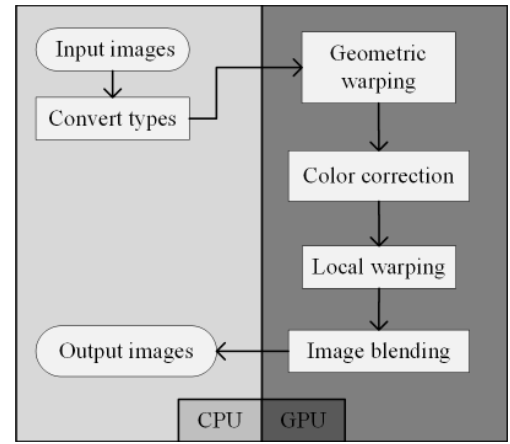

**Fig. 7** The GPU implementation of video stitching.

## 5 Experimental results

### 5.1 Quality evaluation of the stitching system

In this section, we conduct quantitative experiments to evaluate our spatio-temporal color correction algorithm and frame warping algorithm including pre-warping and local warping for video stitching.

**Color correction.** In our paper, we focused on color correction, which helps to get the feature point for our image. The texture is also used to get a better feature. However, the features extracted by texture is a bit redundant. While using feature point can be more efficient and also uniform. We first provide quantitative evaluations of our color correction algorithm. The performance evaluation of color correction is to compute PSNR between the source image and the transferred image, and SSIM [42] which measures the similarity between the target image and the transferred image. Our database sequences have two scenes and each scene contains five frames to evaluate online color correction.

**Table 2** The PSNR comparison. The higher the value of PSNR, the higher similarity the colors are between the two images.

| Scene# / Frame PSNR | Scene# 1 | | Scene# 2 | |
|---|---|---|---|---|
| | 2D-M | Ours | 2D-M | Ours |
| frame = 1 | 18.322 | 21.334 | 20.562 | 23.075 |
| frame = 2 | 18.315 | 21.423 | 20.561 | 23.074 |
| frame = 3 | 18.309 | 21.264 | 20.568 | 23.067 |
| frame = 4 | 18.303 | 21.373 | 20.559 | 23.071 |
| frame = 5 | 18.292 | 21.403 | 20.554 | 23.070 |
| μ | 18.308 | 21.359 | 20.561 | 23.071 |
| σ | 0.0103 | 0.0056 | 0.0045 | 0.0029 |

**Table 3** We compare the SSIM values using our algorithm and Tian's with our data base. The higher the value of SSIM, the less difference between the structure of the two images.

| Scene# / Frame SSIM | Scene# 1 | | Scene# 2 | |
|---|---|---|---|---|
| | 2D-M | Ours | 2D-M | Ours |
| frame = 1 | 0.6724 | 0.7775 | 0.6995 | 0.8062 |
| frame = 2 | 0.6721 | 0.7780 | 0.7016 | 0.8073 |
| frame = 3 | 0.6729 | 0.7777 | 0.7022 | 0.8072 |
| frame = 4 | 0.6744 | 0.7782 | 0.6993 | 0.8074 |
| frame = 5 | 0.6735 | 0.7779 | 0.7002 | 0.8067 |
| μ | 0.6731 | 0.7779 | 0.7006 | 0.8070 |
| σ | 0.0006 | 0.0002 | 0.0004 | 0.0003 |

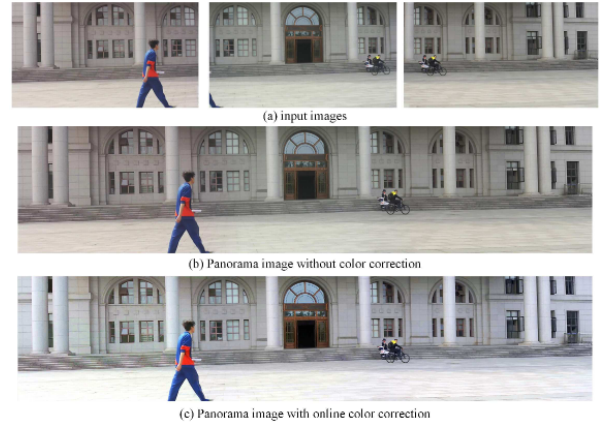

**Fig. 8** The subjective comparison for panorama images with and without online color correction. We use three images to output two panorama images.

For comparison purpose, we test Tian's algorithm[35] under the same database, which is well known as M matrix color correction algorithm. The results are shown in Table 2 and Table 3. From the tables, we can see both the PSNR and SSIM values are improved in our method compared to Tian's method.

Figure 8 compares our panorama image with the stitching results without using color correction. We can observe that the proposed 3D-M color correction can remove the color difference and distortion to produce the seamless panorama image.

We show some exemplar images using our algorithm in figure 9. From figure 9, we can see our algorithm can produce robust and smooth results when images suddenly change in the input.

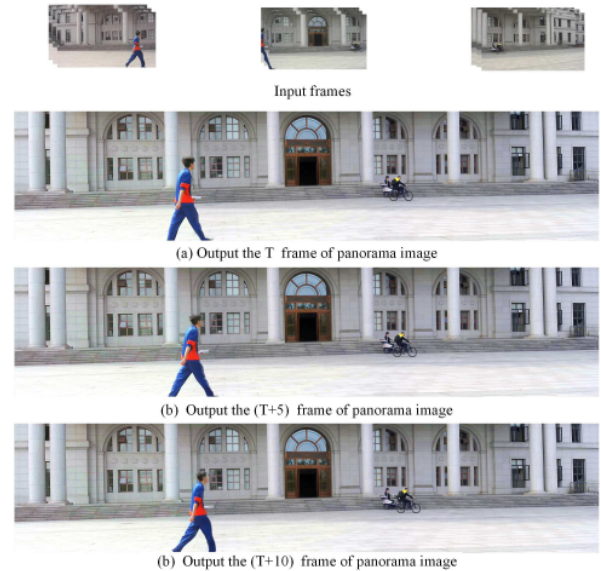

**Fig. 9** The stitched panorama image sequences. We input the source images taken by cameras, and output panorama frames.

**Frame warping**. We utilize our auto-calibration method to calculate accurate homography. We input the initialized frame, then calculate the rough homographic matrixes through rough camera intrinsic parameters.Then, we use the matrixes to split out overlap regions and match SURF feature on the overlap regions. Finally, we calculate the homographic matrixes accurately. We use the homographic matrixes to do the coarse stitching of the input frames. The visualized result is shown in figure 10 (image (a)). Then we use the ho-mographic matrixes to stitch the images in figure 10 (image (b)).

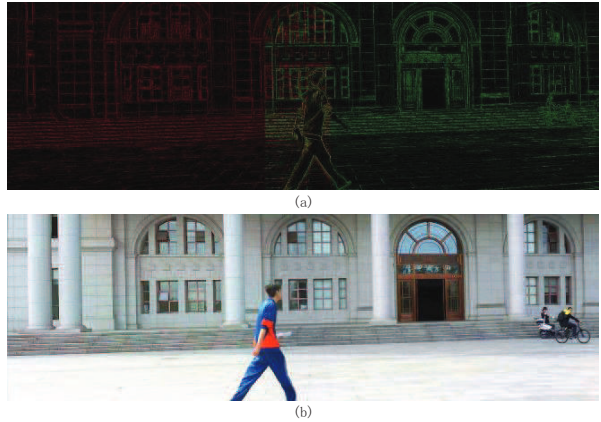

**Fig. 10** Visualized panorama image with simple stitching(a) using auto-calibration and stitching other images using the matrixes calculated by auto calibration(b). The result shows a seamless panorama without ghost, which means our homographic matrixes are accurate.

We evaluate calibration algorithm by using the stitched image between three methods: the baseline AutoStitch in OpenCV for each time instance separately, Photoshop per frame for each time instance separately, and our stitching method using auto stitching and optical flow algorithm. All methods have the same color correction and blending steps. The results are shown from figure 11 to 14. As shown from figure 14, the baseline method has severe ghosting artifacts, while Photoshop per frame reduces some ghosting, but not all. Our results have the least artifact.

### 5.2 Efficiency of the stitching system

We apply GPU implementation for video stitching system and compare the running time. From the results shown in table 4, it is obvious that GPU implementation speeds up the video stitching process. Our system with GPU implementation takes 5 frames per second(FPS) to stitch an image with $4880 \times 1080$ pixels. Moreover, Our system can achieve real-time(50 FPS) with multi-threading and multi-processing technologies.

### 6 Conclusion

We presented an efficient video stitching system with spatio-temporal color correction and robust image warping. Our online color correction uses a novel 3D-M approach for color transfer in the overlap regions. Moreover, we use piecewise function with time information to equalize the color of the global image. To handle inaccurate camera parameters, we presented a semi-auto calibration scheme to adjust the parameters of the skew matrix and the scale matrix to create a new homograph matrix. The video frames are warped in a coarse-to-fine manner. The homography matrix is used to warp images and calculate overlap regions. Then the overlap regions can be finely fused with the optical flow. Experiments show that our GPU accelerated color correction and frame warping can produce plausible panorama video sequences in real time.

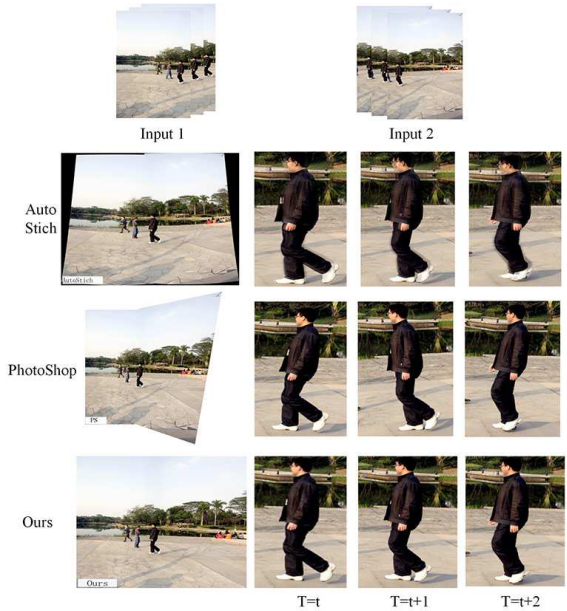

**Fig. 11** Video stitching results of three algorithms on scene 1. We set 2 image sequences as input and the output is a video stitching sequence. Note the ghosting and distortion artifacts on the foreground moving objects.

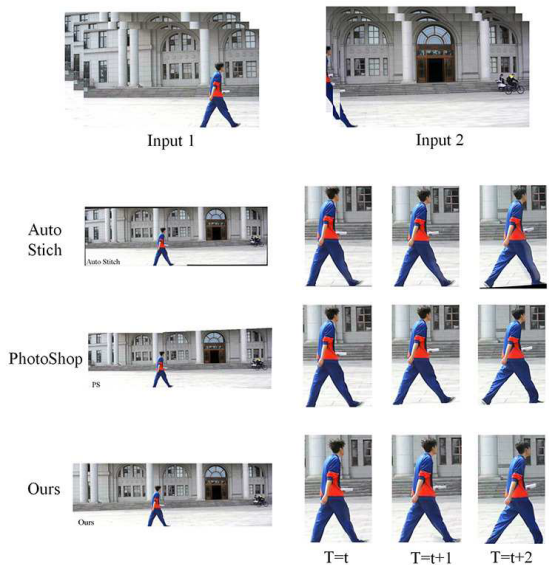

**Fig. 12** Video stitching results of the three algorithms on Scene 2. We set 2 image sequences as input and the output is a video stitching sequence. Note the ghosting and distortion artifacts on the foreground moving objects.

**Table 4** Running time using GPU and single-threaded CPU implementations.

| Scene | Geometric Warping | | Color Correction | | Local Warping | | Image Blending | | All time | | Speed-Up Ratio |
|---|---|---|---|---|---|---|---|---|---|---|---|
| | Time of CPU (s) | Time of GPU(s) | Time of CPU (s) | Time of GPU(s) | Time of CPU (s) | Time of GPU(s) | Time of CPU (s) | Time of GPU(s) | Time of CPU (s) | Time of GPU(s) | |
| #Scene 1 | 3.5 | 0.016 | 3.0 | 0.016 | 48.6 | 0.140 | 5.2 | 0.030 | 59.7 | 0.202 | 284 |
| #Scene 2 | 3.8 | 0.018 | 3.2 | 0.016 | 49.1 | 0.143 | 4.3 | 0.035 | 60.4 | 0.210 | 288 |

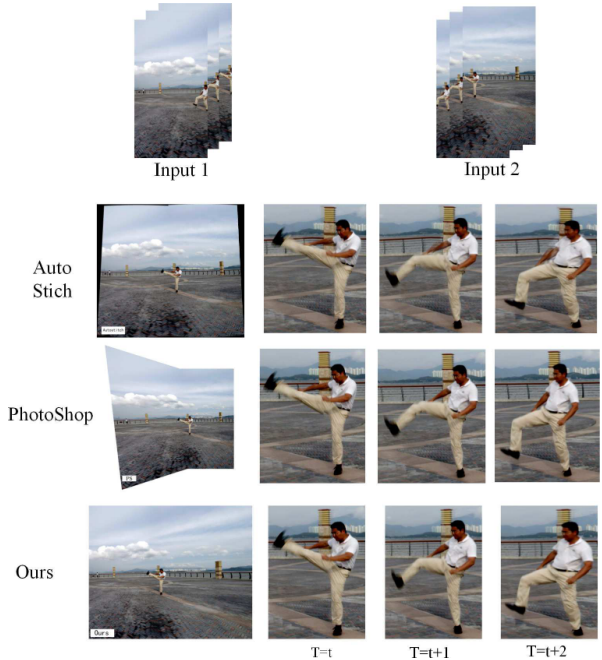

**Fig. 13** Video stitching results of three algorithms on scene 3. We set 2 image sequences as input and the output is a video stitching sequence. Note the ghosting and distortion artifacts on the foreground moving objects.

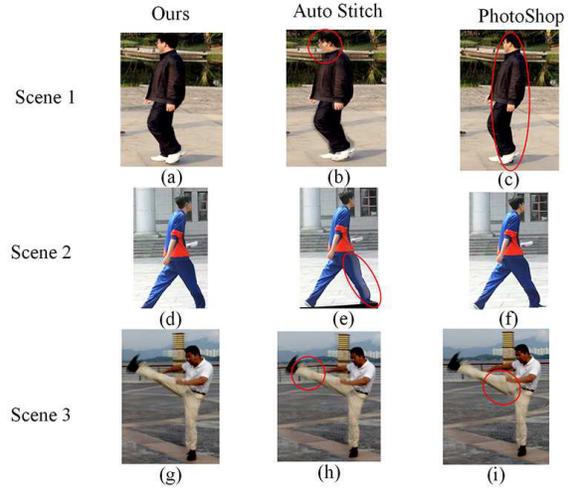

**Fig. 14** We show the stitched overlap regions with people of the video stitching results from the three algorithms.